\documentclass{article}

\usepackage{PRIMEarxiv}
\usepackage{multirow}
\usepackage{makecell}
\usepackage{amsmath}
\usepackage{afterpage}
\usepackage{rotating}

\usepackage[utf8]{inputenc} % allow utf-8 input
\usepackage[T1]{fontenc}    % use 8-bit T1 fonts
\usepackage{hyperref}       % hyperlinks
\usepackage{url}            % simple URL typesetting
\usepackage{booktabs}       % professional-quality tables
\usepackage{amsfonts}       % blackboard math symbols
\usepackage{nicefrac}       % compact symbols for 1/2, etc.
\usepackage{microtype}      % microtypography
\usepackage{lipsum}
\usepackage{graphicx}
\graphicspath{{media/}}     % organize your images and other figures under media/ folder
\usepackage{float}
  
\title{Rare Class Prediction Model for Smart Industry in Semiconductor Manufacturing}

\author{
  Abdelrahman Farrag, Mohammed-Khalil Ghali, Yu Jin\\
  School of Systems Science and Industrial Engineering \\
  State University of New York at Binghamton \\
  Binghamton, NY, USA\\
  \texttt{\{afarrag1, mghali1, yjin\}@binghamton.edu} \\
}

\begin{document}
\maketitle

\begin{abstract}
The evolution of industry has enabled the integration of physical and digital systems, facilitating the collection of extensive data on manufacturing processes. This integration provides a reliable solution for improving process quality and managing equipment health. However, data collected from real manufacturing processes often exhibit challenging properties, such as severe class imbalance, high rates of missing values, and noisy features, which hinder effective machine learning implementation. In this study, a rare class prediction approach is developed for in situ data collected from a smart semiconductor manufacturing process. The primary objective is to build a model that addresses issues of noise and class imbalance, enhancing class separation. The developed approach demonstrated promising results compared to existing literature, which would allow the prediction of new observations that could give insights into future maintenance plans and production quality. The model was evaluated using various performance metrics, with ROC curves showing an AUC of 0.95, a precision of 0.66, and a recall of 0.96.
\end{abstract}


\section{Introduction}
Manufacturing of semiconductor wafers involves hundreds of sophisticated fabrication processes, many of which are performed repeatedly, such as oxidation, photolithography, cleaning, etching, and planarization. Wafer yield is calculated as the ratio of the total conforming products to the total semiconductor chips within the wafer. Maintaining a high yield through reliable and accurate quality control is crucial for success in the semiconductor industry \cite{kumar2006review}. A key step for yield improvement is identifying the operations that significantly impact wafer yield, known as “critical process steps.”

Selecting critical process steps poses significant challenges due to the inherent complexities of process data. These data are predominantly derived from numerous in-situ sensors, introducing high dimensionality with potential noisy features—a common characteristic in operational datasets. Additionally, the data suffer from a high rate of missing values, primarily due to low measurement frequencies and the limitations of current measurement technologies. During fabrication, each wafer undergoes various process steps and is subsequently inspected by measurement equipment. Due to the time-intensive nature of these inspections and the limited capacity of the measurement tools, only a small fraction of the wafers are actually measured. This practice of random sampling for measurement further complicates data analysis \cite{shehata2018implementation, shehata2019identifying, shehata2022reduction}. For instance, if there are five process steps and the measurement rate is only 20\%, with wafers being randomly and independently selected for inspection, the likelihood of obtaining complete measurement data across all steps is 0.032\% \cite{lee2019data}. This issue is magnified in practice, where over 500 process steps might exist, making it challenging to establish correlations among process steps. Furthermore, the majority of the wafer fabrication lines, as they mature, yield a high number of wafers with acceptable quality, reducing the incidence of low-yield wafers. However, to effectively investigate and improve wafer yield, it is equally important to analyze both high and low-yield wafers. The scarcity of low-yield wafers makes it difficult to assess the full impact of process variability on overall production quality.

With the advent of Industry 4.0, the integration of advanced machine learning (ML) techniques with predictive maintenance regimes has become a cornerstone for enhancing operational efficiencies in production lines. These methodologies are critical for the early detection of anomalies and the prevention of mechanical failures, thereby streamlining production processes, reducing costs, and ensuring safety. However, the efficacy of these models hinges on the proper selection of the model variables or the production operation features. Therefore, to address the challenges associated with semiconductor wafer fabrication, we propose a structured, data-driven approach to identify critical process steps through a voting-based feature selection approach. This approach aims to significantly select the most contributing features to the dependent variable. Moreover, our developed approach addresses the multifaceted challenges associated with in-situ data in semiconductor manufacturing, such as data imbalance, missing entries, and the presence of outliers. By incorporating an optimized strategy for imputing missing data, our methodology ensures imputing missing data, thereby aligning the dataset more closely with its inherent distribution.

The remainder of this paper is organized as follows. We begin with a literature review considering imputation strategies, high-class imbalance challenges, and feature selection methods. Following this, the methodology section outlines the proposed approach, detailing the data preprocessing techniques and the specifics of the case study. Subsequent sections present the results of the model and discuss their implications. Finally, the paper concludes with a summary of the findings and suggests directions for future work

\section{Literature review}
The literature review is structured into three primary sections, each focusing on a critical aspect of data preprocessing and feature selection in the context of semiconductor manufacturing. Firstly, the review addresses the prevalent issue of missing data in the dataset. Various imputation strategies are explored to manage and rectify the gaps in data. Secondly, the review delves into the challenge of class imbalance, a common issue in predictive modeling. Different resampling techniques are reviewed for their effectiveness in balancing the dataset. Finally, feature selection methods are discussed to identify the most influential features for the classification model, aiming to enhance the predictive accuracy and efficiency of the model.

\subsection{Data imputation methods}
The issue of missing data presents a common and significant challenge in numerous studies, impacting the reliability of statistical analyses through potential information loss and biases in parameter estimation \cite{rubin2018multiple}. Missing data are classified into three formalized mechanisms: missing completely at random (MCAR), missing at random (MAR), and missing not at random (MNAR) \cite{rubin1976inference}. MCAR occurs when the absence of data is independent of any observed or unobserved variables, suggesting no systematic loss. In contrast, MAR involves missing instances related to other measured variables but not to the missing values themselves, indicating a systematic relationship influenced by other dataset variables \cite{baraldi2010introduction}. MNAR, the most complex, occurs when missing data depend on the missing values themselves. Identifying the precise mechanism in practical scenarios, such as semiconductor manufacturing where wafers are randomly selected for measurement, can be challenging. In such contexts, the missing data are likely MAR, correlating with observed values due to interconnected process steps, rather than MNAR.

Traditional imputation methods, such as deletion and mean imputation, are effective primarily when data are MCAR. Conversely, modern techniques like maximum likelihood, multiple imputation, hot-deck imputation, regression imputation, expectation maximization (EM) \cite{dempster1977maximum}, and Markov chain Monte Carlo (MCMC) are designed to provide unbiased estimates for data classified as either MCAR or MAR. The prevalence of missing data significantly affects the quality of statistical inferences, although there is no universally accepted threshold for an acceptable percentage of missing data. It has been suggested that a missing rate of 5\% or less is generally negligible, whereas rates exceeding 10\% are likely to introduce bias into the statistical analysis \cite{bennett2001can, dong2013principled, schafer1999multiple}. 

A novel data imputation approach \cite{salem2018experimental}, known as "In-painting KNN-imputation", is developed and compared with the mean imputation strategy after applying various ML approaches. The developed approach outperformed the common data imputation technique, the mean imputation. The performance metrics are significantly improved after applying the In-painting KNN imputation such that the recall percentage increased by 10\% and the AUC increased by 5\%. In \cite{jelinek2015diagnostic}, authors applied discretization of continuous features to make all data nominal, and it is effective in missing values imputing; especially, no unique approach is required for different feature types. This approach was compared by applying prediction classification models on the SECOM dataset \cite{misc_secom_179} after and before the imputation approach, which showed significant improvement. In \cite{lee2019data}, a systematic and data-driven approach was proposed for identifying the critical process steps through missing value imputation using an EM algorithm, data resampling using SMOTE, and feature selection using MeanDiff. LR, k-nearest Neighbor (KNN), and Support Vector Machines (SVM) are applied and compared.

\subsection{Resampling methods}
Defective data sampling in the context of ML and data analysis is a critical issue, particularly when dealing with datasets related to quality control or failure detection. In these scenarios, the data is often imbalanced, characterized by a significant discrepancy between the 'defective' or 'positive' class (such as instances of failure or defect) and the 'non-defective' or 'negative' class. This imbalance poses substantial challenges for predictive modeling, as the rarity of the defective class compared to the non-defective class can lead to models that are biased and ineffective in accurately identifying defects. The model, skewed towards the majority class, might exhibit high accuracy while failing to effectively identify instances of the minority class, resulting in an increased rate of false negatives \cite{thabtah2020data}. This is particularly concerning in defect detection where missing actual defects (false negatives) can have serious implications. Moreover, the imbalance creates a challenging precision-recall trade-off, where improving one often compromises the other. To address these issues, techniques such as resampling the data (either oversampling the minority class or undersampling the majority class), using different performance metrics (like F1-score, precision-recall curves, ROC-AUC), and employing algorithms that are specifically designed for imbalanced data can be effective.

Undersampling techniques in machine learning effectively address class imbalance by retaining the most representative instances of the majority class, thereby optimizing the learning process \cite{mohammed2020machine}. Integrating data-driven models with undersampling methods significantly advances this approach, selectively undersampling instances near the minority class to better mitigate imbalance issues \cite{arefeen2020neural}. Specific methods like Cluster-Based, Tomek Link, and Condensed Nearest Neighbours refine decision boundaries, enhancing classifier accuracy \cite{bansal2021analysis, pereira2020mltl}. Each undersampling technique, including Edited Nearest Neighbors (ENN), Near Miss, and the Neighborhood Cleaning Rule (NCR), offers unique benefits and challenges. ENN uses the k-nearest neighbor algorithm to remove noisy majority class instances but can be computationally intensive and risk information loss \cite{wang2023synthetic}. Near Miss prioritizes majority class instances close to the minority class, which can introduce bias by excluding distant but important instances \cite{tanimoto2022improving}. NCR balances cleaning ambiguous majority class examples and preserving minority class integrity, although it may over-clean and exclude valuable information \cite{laurikkala2001improving, van2007experimental}.

On the other hand, oversampling techniques address class imbalance by augmenting the minority class. Comparative studies between undersampling and oversampling highlight the importance of choosing the right method, particularly in fields requiring high prediction accuracy \cite{jeong2022comparative}. Random Oversampling duplicates minority class instances, which can lead to overfitting \cite{zheng2015oversampling}. Methods like SMOTE (Synthetic Minority Over-sampling Technique) generate synthetic instances to enhance diversity but can introduce noise \cite{chawla2002smote}. Borderline-SMOTE creates samples near decision boundaries, useful for closely spaced classes \cite{han2005borderline}. ADASYN (Adaptive Synthetic Sampling) targets harder-to-learn minority instances but also risks noise introduction \cite{he2008adasyn}. Each oversampling technique must balance augmenting the minority class and maintaining data quality, with trade-offs between representation and potential noise.

\subsection{Feature selection methods}

Feature selection algorithms such as Boruta, Multivariate Adaptive Regression Spline (MARS), and the Principal Component Analysis (PCA) were applied to select the most important features. The results showed better values for precision when the features were selected using Boruta and MARS rather than PCA and better values for accuracy when the data was unbalanced and classified using Random Forest (RF) and Logistic Regression (LR) rather than Gradient-Boosted Trees (GBT).  In \cite{kerdprasop2010feature}, several approaches for feature selection were used, such as Chi-Square, mutual information, and PCA. Classification models such as LR, KNN, Decision Trees (DT), and Naïve Bayes (NB) were applied, where DT showed the best results compared to others with an F-measure of 64\% and precision of 67\%. In \cite{thabtah2020data}, the high-class imbalance problem was considered using SMOTE, and PCA was applied to reduce the high dimensionality. Receiver Operating Characteristic (ROC) evaluated the models, which showed the RF has the best AUC of 0.77 compared to KNN and LR. In \cite{munirathinam2016predictive}, a decision model was developed to detect equipment failure as soon as possible to maintain high productivity and efficiency. Four prediction models were carried out after data preprocessing and feature selection, and NB showed the best results compared to KNN, DT, SVM, and ANN. On the same track, \cite{carbery2019new} aimed at improving the accuracy of the classification prediction model of the SECOM dataset. They employed an early detection prediction model using the framework published in \cite{carbery2018new} and used XGBoost, which showed significant results compared to RF and DT. Another major research challenge related to applying deep learning and optimizing the classification performance using metaheuristic approaches \cite{deb2020recent} to improve the predictability of the SECOM dataset. In \cite{moldovan2018chicken}, an approach was proposed by applying neural networks and using the Chicken Swarm Optimization (CSO) algorithm to optimize the hidden layer nodes. The approach showed an accuracy of 70\%, recall of 65\%, and precision of 73\%. Another approach used by \cite{moldovan2020particle}, an ensemble of deep learning models, was applied, and particle swarm optimization (PSO) was used to determine the model weights. The approach is evaluated and shows better results compared to KNN, RF, AdaBoost, and GBT. 

Most of the classification models are developed based on accuracy; however, these prediction models present a paradox in terms of accuracy. Accuracy paradoxes do not emphasize the prediction of all classes but rather on the prediction of the majority of classes. In data imbalance concerns, accuracy alone is not sufficient. Predicting a minority class is difficult in rare class imbalance datasets because the rare class is small compared to the majority class. Because majority class prediction is simple, its accuracy is easily classified. Minority classes, on the other hand, are difficult. As a result, when the performance of a prediction model is measured just by its accuracy, the minority classes cannot be predicted. As a result, even if the degree of accuracy is excellent, there is a significant likelihood of predicting only the majority classes without considering the minority classes. Balanced accuracy is a crucial evaluation performance metric for these cases of highly imbalanced data. Some of the previous research employed sampling strategies to increase the number of minority classes. However, suppose the features are chosen based on the data distribution, before the oversampling of the minority class or the undersampling of the majority class.

\section{Methodology}
This methodology section outlines our approach to addressing the challenges associated with in-situ sensor data in semiconductor manufacturing. It includes the case study applied and the array of data preprocessing techniques employed. These preprocessing steps encompass dealing with missing values, splitting the dataset, and scaling the data. Additionally, the methodology involves feature selection to isolate variables and data resampling techniques aimed at correcting class imbalances. 

\begin{figure}[H]
    \centering
    \includegraphics[width=1.0\textwidth]{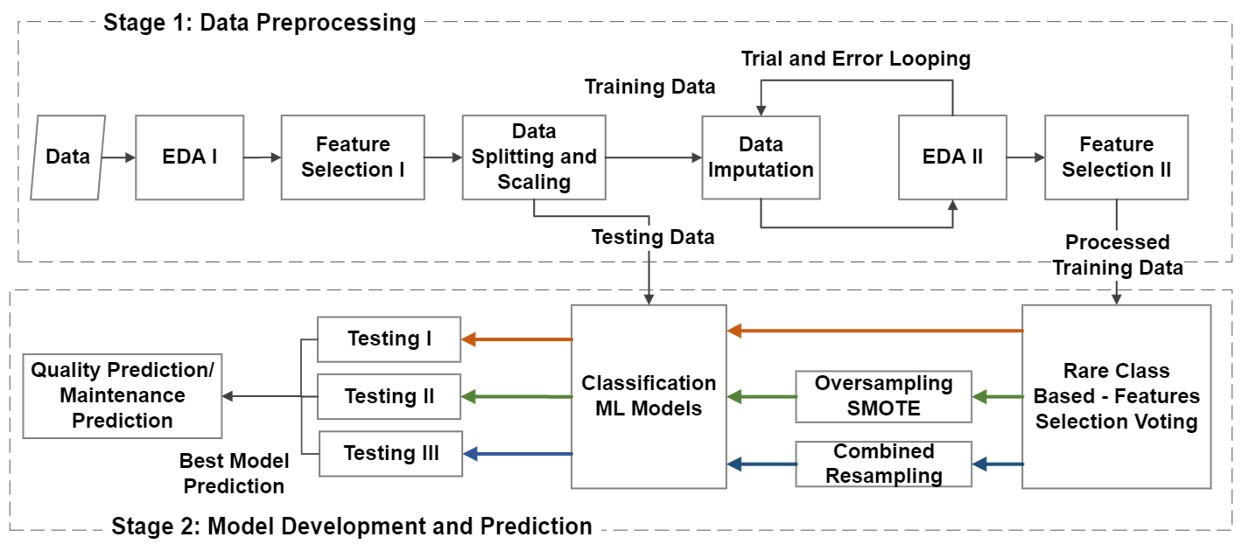}
    \caption{A schematic overview for the proposed approach.}
    \label{fig 1}
\end{figure}

\subsection{Proposed approach}
The proposed methodology, as illustrated in Figure \ref{fig 1}, is structured into two main stages: Data Preprocessing and Model Development and Prediction. The process begins with an initial Exploratory Data Analysis I (EDA I), where preliminary insights into the data are gathered. This is followed by the first phase of feature selection, aimed at reducing dimensionality early in the process. Subsequently, the data is split and scaled to prepare for imputation, where missing values are filled to ensure data completeness. The processed data then undergoes an EDA II in a trial and error-loop, allowing for further refinement and understanding. This leads to a second phase of feature selection (Feature Selection II), where more focused and definitive feature choices are made based on the insights gained. The outcome is a set of processed training data, ready for the next stage. In the Model Development and Prediction stage, the training data is subjected to multiple tests through various ML classification models. Techniques such as SMOTE and modified SMOTE are applied to address class imbalance by resampling the data, particularly enhancing the representation of rare classes. Feature selection is further refined through a voting mechanism based solely on features from the minor class, ensuring that only the most relevant attributes are used in the final model. The methodology supports multiple testing scenarios (Testing I, II, and III), each providing a different perspective on model performance and robustness. Ultimately, the best-performing model is selected for quality prediction or maintenance prediction applications, focusing on forecasting and operational improvements.

\subsection{Case Study}
\begin{figure}[H]
    \centering
    \includegraphics[width=1.0\textwidth]{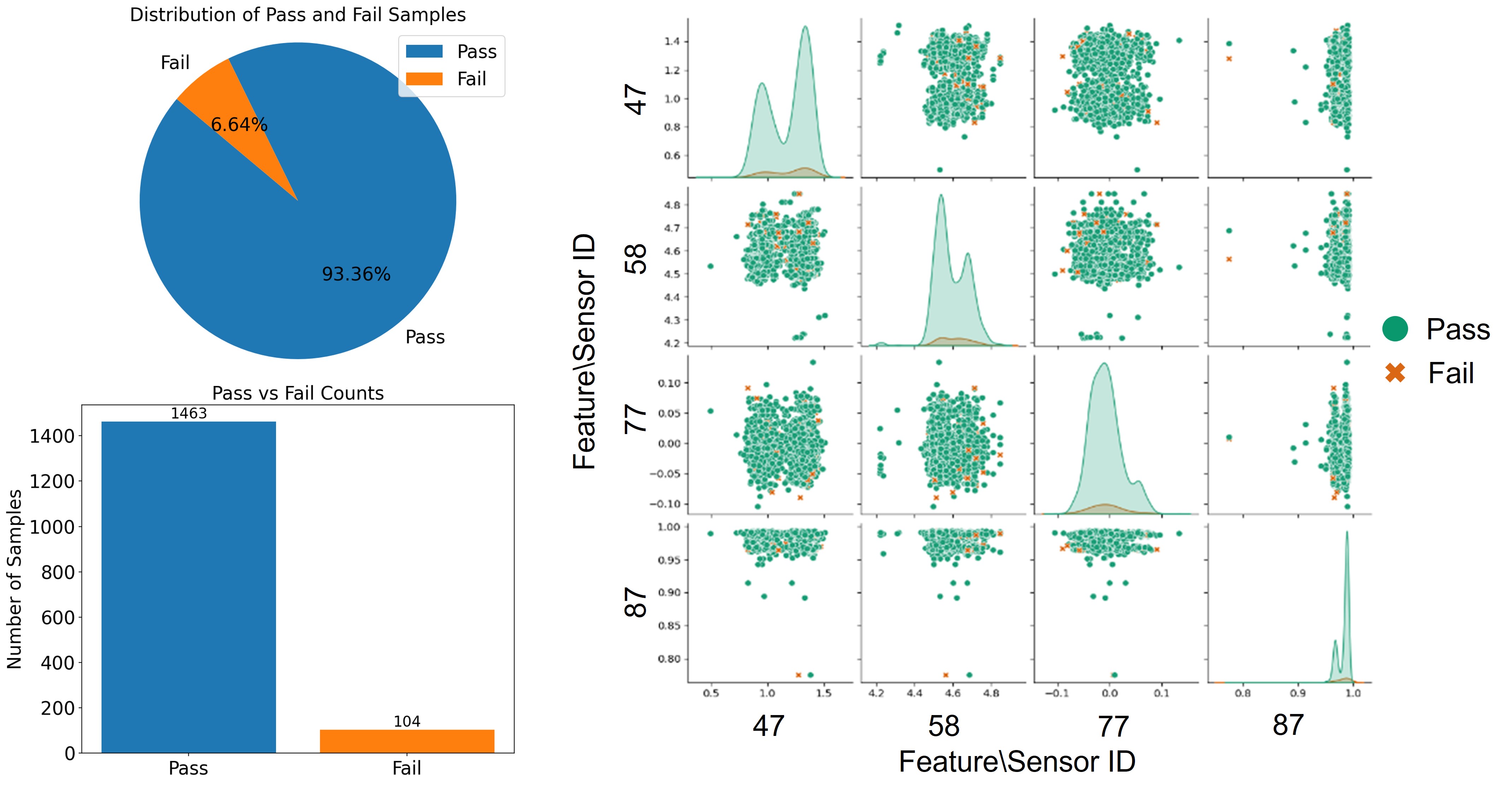}
    \caption{Exploratory Data Analysis for SECOM Data.}
    \label{fig 2}
\end{figure}

\begin{figure}[H]
    \centering
    \includegraphics[width=1.0\textwidth]{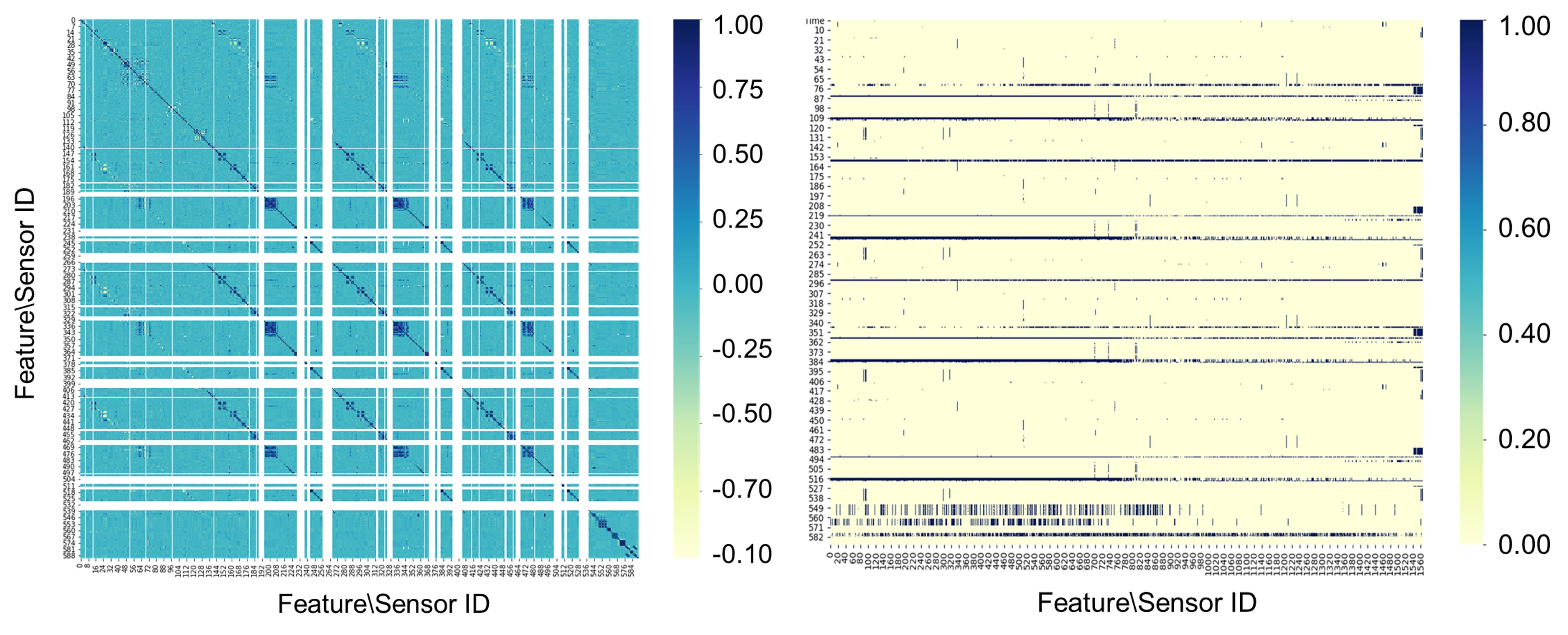}
    \caption{Features analysis for SECOM Data.}
    \label{fig 3}
\end{figure}

This study uses the SECOM dataset, an open-source industrial dataset representative of complex semiconductor manufacturing processes \cite{misc_secom_179}. Handling semiconductor data presents multiple challenges. The dataset comprises 591 sensor measurements across 1567 samples, with only 104 samples identified and classified as failures. Given the high costs associated with semiconductor manufacturing, processes are controlled to minimize defects, resulting in a pronounced class imbalance in our dataset at a ratio of 1:14, as shown in Figure \ref{fig 2}. Additionally, as a reflection of real-world engineering environments, the dataset is prone to missing data attributed to sensor faults or operational oversights, as shown in Figure \ref{fig 3}. Specifically, it exhibits a significant level of missing data, at 4.5\%, with 28 sensor measurements frequently unreported. Moreover, many of the sensor measurements are not directly correlated with the final classification target, introducing noisy features that complicate the analysis.

The SECOM dataset has been extensively utilized by researchers aiming to address real-world classification challenges such as fault diagnosis and detection. Scholarly investigations into this dataset typically focus on three main areas: addressing the high volume of missing data through various imputation strategies, managing the pronounced class imbalance with different resampling techniques, and optimizing feature selection to enhance model accuracy. These studies collectively aim to develop an optimal classification model that effectively navigates the inherent complexities of semiconductor manufacturing data. Despite advancements, current approaches to these data challenges remain insufficient. The complexity and variability in semiconductor manufacturing mean imputation strategies often miss key data patterns. Resampling techniques may not fully address class imbalances, leading to biased models. Feature selection methods, while improving accuracy, struggle to identify critical features in noisy datasets. Thus, more robust and comprehensive methods are needed to better handle these intricacies and ensure reliable, accurate predictive models.

\subsection{Data preprocessing}

Depending on the dataset features, data preprocessing for minor class mining employs various strategies. It is crucial to mix several strategies and apply them in the proper sequence to accomplish minor class prediction. The runs were carried out by separating the training sets (70\%) and testing sets (30\%).

\subsubsection{Missing Values}

EDA is initially performed to understand the behavior of the data and identify missing values. The percentage of missing values is calculated, and features with a missing value percentage greater than 50\% are removed, along with features that have constant values, as they do not contribute to the dependent variable. EDA is then performed again to analyze the data distribution for each feature and the overlap between classes for each pair of features. This step is crucial for selecting the proper missing data imputation strategy. 

Several imputation strategies are considered, including forward imputation, backward imputation, mean imputation, most frequent imputation, linear interpolation, k-Nearest Neighbors (k-NN) imputation, and Multivariate Imputation by Chained Equations (MICE) \cite{azur2011multiple}. 

k-NN imputation involves filling in missing values using the most similar instances from the dataset. For a given missing value $x_i$, the imputation is performed using the mean of the $k$ nearest neighbors:

\begin{equation}
\hat{x}_i = \frac{1}{k} \sum_{j=1}^{k} x_j
\end{equation}

where $x_j$ represents the $j$-th nearest neighbor to the instance with the missing value $x_i$.

MICE handles multivariate missing data by iteratively imputing each variable using a chained equation approach. Each incomplete variable is modeled conditionally using the other variables in the dataset. The imputation process can be summarized as follows:

1. Initialize missing values with initial guesses (e.g., mean or median).
2. For each variable with missing data, $\mathbf{X}j'$:
    \begin{itemize}
        \item Regress $\mathbf{X}_j$ on the other variables $\mathbf{X}_{j'}$.
        \item Draw imputed values for the missing entries in $\mathbf{X}_j$ from the predictive distribution obtained in the regression.
    \end{itemize}
3. Repeat step 2 for a specified number of iterations.

The general form of the regression model used for imputation is:

\begin{equation}
\mathbf{X}_j = \mathbf{X}_{j'} \beta_j + \epsilon_j
\end{equation}

where $\mathbf{X}_{j'}$ represents the other variables, $\beta_j$ is the regression coefficient, and $\epsilon_j$ is the error term.

Then, a loop is performed until the best class separation and distribution are achieved. For features with skewed data distribution, median imputation is applied. For those with a Gaussian mixture distribution, mean imputation is used.

\subsubsection{Data Splitting}

Stratified cross-validation is used to split the data, which is particularly useful for imbalanced datasets. A portion of the data (training set) is used to train the algorithms, while the remainder (testing set) is used to estimate the algorithms' performance. A 5-fold cross-validation technique is employed, where the data is randomly divided into five subgroups with an equal number of samples. The processes described in the following subsections are performed five times, with one fold used as testing data and the remaining four folds used as training data. The obtained model is tested using the testing data and evaluated using performance metrics.

\subsubsection{Data Scaling}
Due to the irregular state of the data, scaling is required to normalize the dataset. Feature scaling can enhance the classification performance of learning algorithms. The data is normalized to a linear scale from 0 to 1 using the following equation:

\begin{equation}
\hat{X} = 0.5 + \frac{X - \text{Ave}(X)}{\text{Max}(X) - \text{Min}(X)}
\end{equation}

where $\text{Min}(X)$ is the minimum value of the data, $\text{Max}(X)$ is the maximum value of the data, and $\text{Ave}(X)$ is the average value of the data.

\subsection{Feature Selection}

Given the hundreds of features, most of which are unnecessary, feature selection is crucial for rare class prediction to create an effective prediction model. The developed model is biased towards the rare class features, giving priority to features that contribute significantly to the rare class. The selection of features is a critical step in these types of problems, as the selection algorithm might be affected by the high dimensionality of the features. Therefore, a voting strategy is employed, selecting features that have been chosen by three or more of the feature selection methods, in addition to those contributing to the minor class. 

The feature selection methods used in this process are ANOVA/F-value, mutual information, Boruta algorithm, MARS, Recursive Feature Elimination (RFE) by Logistic Regression (LR), RFE by Support Vector Machine (SVM), RFE by Random Forest (RF), LASSO, and Sequential Feature Selection (SFS) by XGBoost and SVM. RFE is a technique that recursively removes the least important features and builds the model with the remaining features. SFS is employed, which incrementally builds or reduces the feature set by adding or removing features based on their impact on the model's performance. Therefore, twelve feature selection methods participated in the voting process, with the criteria for selection being features that received votes from at least three methods. This process is repeated until the optimal number of features is reached. The feature selection voting results showed that 21 features were disregarded by the voters, while 183 features were voted for by the feature selection methods. However, only two features were selected by all 12-feature selection methods. In our case, we ultimately selected 81 features.

\subsection{Data Resampling}

The main aim of data resampling is to solve the data imbalance problem between the minority and majority classes. This step is carried out only for the training dataset to prevent overfitting of the testing data. Two different strategies are performed: oversampling of the minority class using the Synthetic Minority Oversampling Technique (SMOTE) and under-sampling of the majority class combined with SMOTE. 

SMOTE is applied to the minority class by 70\%, creating synthetic data points by interpolating between existing data points. The new synthetic data point $\mathbf{x}_{new}$ is generated using the formula:

\begin{equation}
\mathbf{x}_{new} = \mathbf{x}_i + \lambda (\mathbf{x}_j - \mathbf{x}_i)
\end{equation}

where $\mathbf{x}_i$ and $\mathbf{x}_j$ are existing minority class instances, and $\lambda$ is a random number between 0 and 1.

The combined under-sampling and SMOTE strategy involves oversampling the minority class by 40\% and under-sampling the majority class by 80\%, adjusting the ratio from 1:14 to approximately 4:5. In both resampling approaches, efforts are made to make the classes closer in size. This prevents half of the model data from being synthetic due to the large initial class imbalance. These methods collectively aim to address the issues of class imbalance and ensure that the model can generalize better to unseen data.

\subsection{Performance Metrics}

Several metrics are used for the evaluation of the testing data results. Balanced accuracy is particularly important for unbalanced and rare class data because it accounts for the imbalance by averaging sensitivity and specificity. It ensures that both classes are equally considered in the performance evaluation, providing a more comprehensive measure of model performance. It is calculated as:
\begin{equation}
\text{Balanced Accuracy} = \frac{\text{Sensitivity} + \text{Specificity}}{2}
\end{equation}

Precision measures the accuracy of the positive predictions, indicating the proportion of true positive results among all positive predictions. It is defined as:
\begin{equation}
\text{Precision} = \frac{\text{True Positives}}{\text{True Positives} + \text{False Positives}}
\end{equation}

Recall, also known as sensitivity, measures the ability of the model to identify all relevant instances, indicating the proportion of true positive results among all actual positive instances. It is calculated as:
\begin{equation}
\text{Recall} = \frac{\text{True Positives}}{\text{True Positives} + \text{False Negatives}}
\end{equation}

False Alarm Rate (FAR) measures the proportion of false positive results among all negative instances. It is given by:
\begin{equation}
\text{FAR} = \frac{\text{False Positives}}{\text{False Positives} + \text{True Negatives}}
\end{equation}

The Receiver Operating Characteristic (ROC) curve is an evaluation metric for binary classification problems. It is a probability curve that plots the True Positive Rate (TPR) against the False Positive Rate (FPR) at various threshold values, effectively separating the ‘signal’ from the ‘noise’. The Area Under the Curve (AUC) is a measure of the classifier's ability to distinguish between classes and serves as a summary of the ROC curve. A higher AUC indicates better model performance.

\section{Results}
\subsection{Data pre-processing}

Firstly, a random feature pair plot is performed for the data, as shown in Figure \ref{fig 2}. It is observed that the data classes are completely overlapped and irregularly distributed. The percentage of missing values is estimated and found to be 4.5\%, resulting in the removal of 28 columns with a missing values rate greater than 50\%. Six different imputation approaches are adopted for the remaining 1.26\% missing values. k-NN has shown the best separation of data after applying it; however, some features are imputed using the median and others using the mean to properly fit normal distribution curves. 

\begin{figure}[H]
    \centering
    \includegraphics[width=0.6\textwidth]{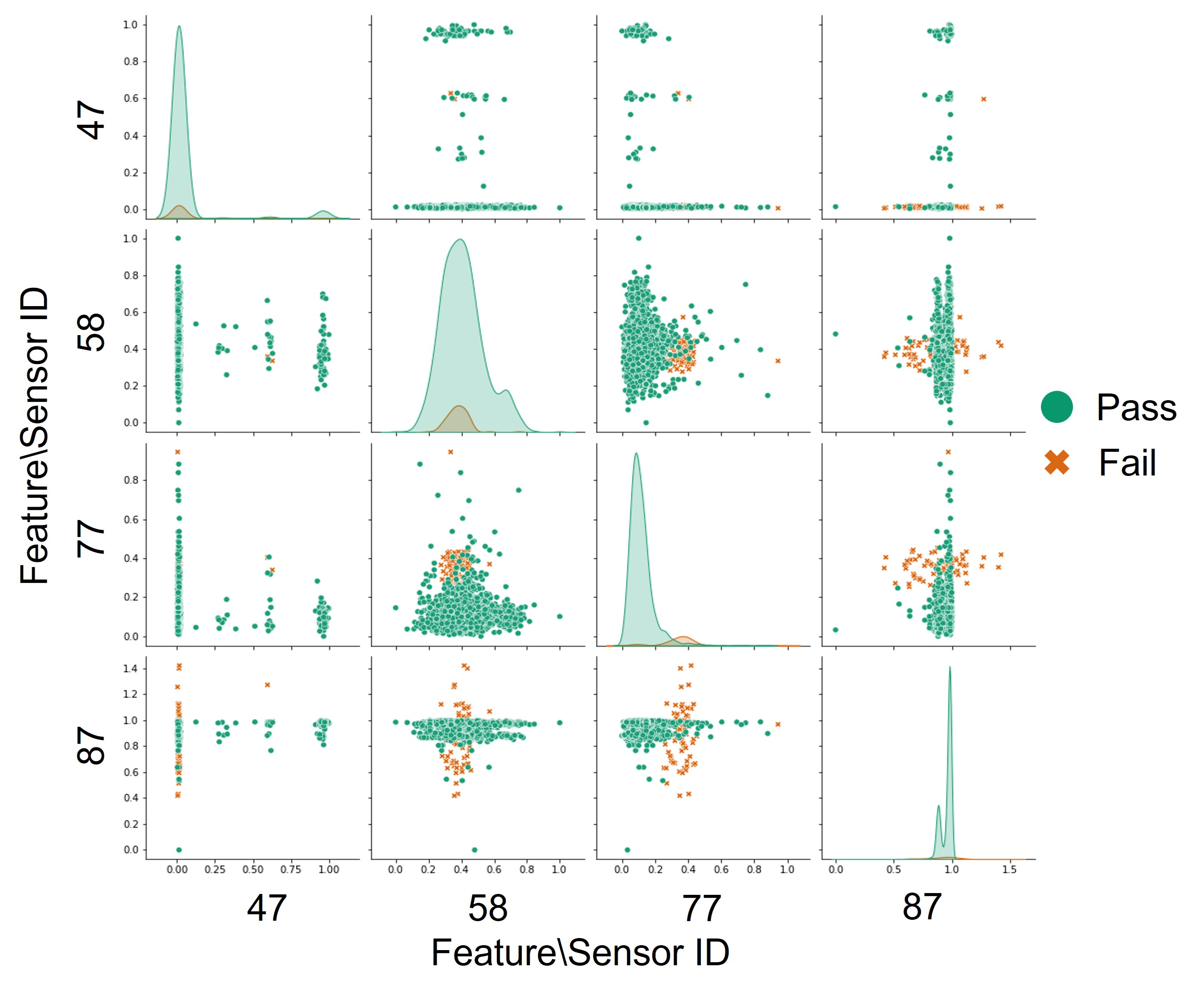}
    \caption{EDA for SECOM Data after data pre-processing.}
    \label{fig 4}
\end{figure}

A heat map is generated to show the correlation of features, revealing that 116 features have constant values. Features with a correlation greater than 70\% are removed, resulting in 204 features. After these steps, 1567 observations are ready for splitting. The data is then split, with 30\% reserved for testing. The training data are normalized, and pair plotting is performed, as shown in Figure \ref{fig 4}. Figure It shows an improvement in the separation and distribution of the training data, which will aid in the effective training of the developed model. However, it is noted that the data is still imbalanced.

\subsection{Rare class-based features selection voting}
\begin{figure}[H]
    \centering
    \includegraphics[width=1.0\textwidth]{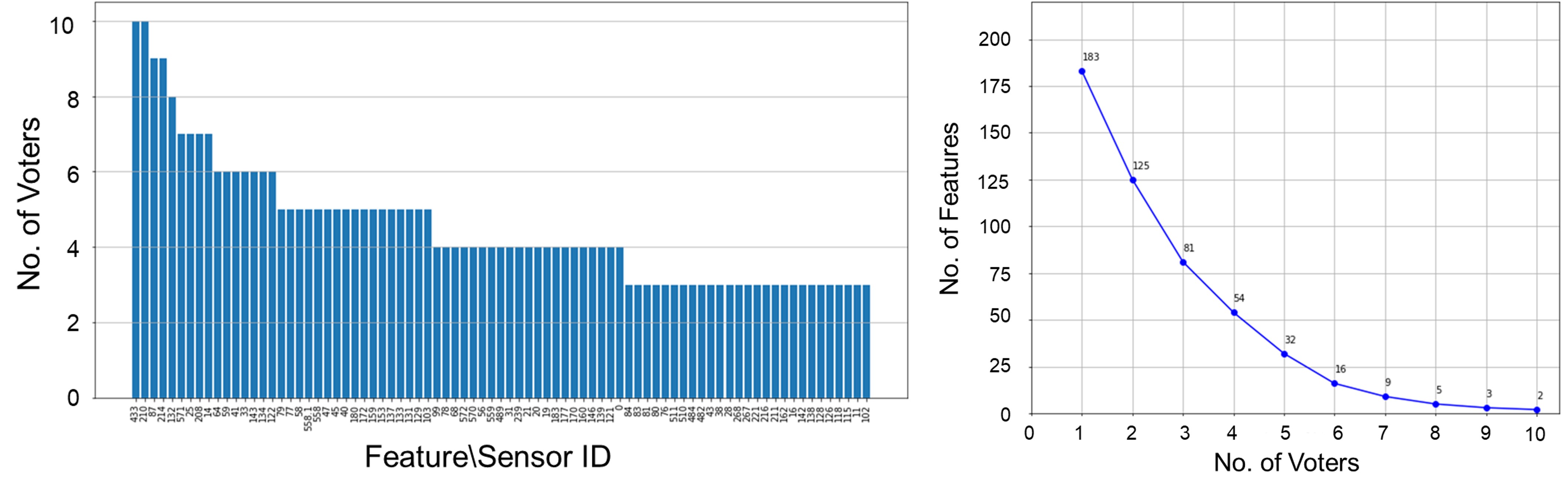}
    \caption{Voting results of the rare class-based features selection approach.}
    \label{fig 5}
\end{figure}

The feature selection approach has resulted in 183 features being voted for inclusion, as shown in Figure 3. A threshold of at least three votes for each feature is considered, resulting in 81 features being selected for subsequent steps. The selected features, with voting greater than or equal to 3, are shown in Figure \ref{fig 5} in descending order, with features 433 and 210 being voted for by all feature selection algorithms.

The threshold of 3 votes was chosen because the model showed more stability at this level. Different feature selection algorithms tended to generate varying numbers of features with each run. Therefore, to achieve more consistent and intuitive results, the different algorithms assume that a feature should be selected at least three times.

\subsection{Classification prediction evaluation}
This section presents the results for classification models across three runs. Each run's results are demonstrated with a table of performance metrics and ROC curves to show the AUC. Finally, a summary plot for the performance metrics of the three runs is performed.

\subsubsection{Testing Scenario I: Imbalanced models}

The results of the first run show that XGB and DTC have the best performance metrics, closely followed by GBC, which shows a relatively low precision value. However, LR, SVM, and RF did not perform well. Despite the high precision value of RF at 100\%, it does not necessarily mean it could predict all positives correctly. It indicates that the RF model predicts all the negative cases but fails to detect the positive cases since the recall is considerably low at 17\%. The best model should have the highest precision, recall, balanced accuracy, AUC, and the lowest false alarm rate. It can be observed that XGB has relatively the highest values and is the best model in terms of imbalanced data.

\begin{figure}[h]
    \centering
    \includegraphics[width=0.5\textwidth]{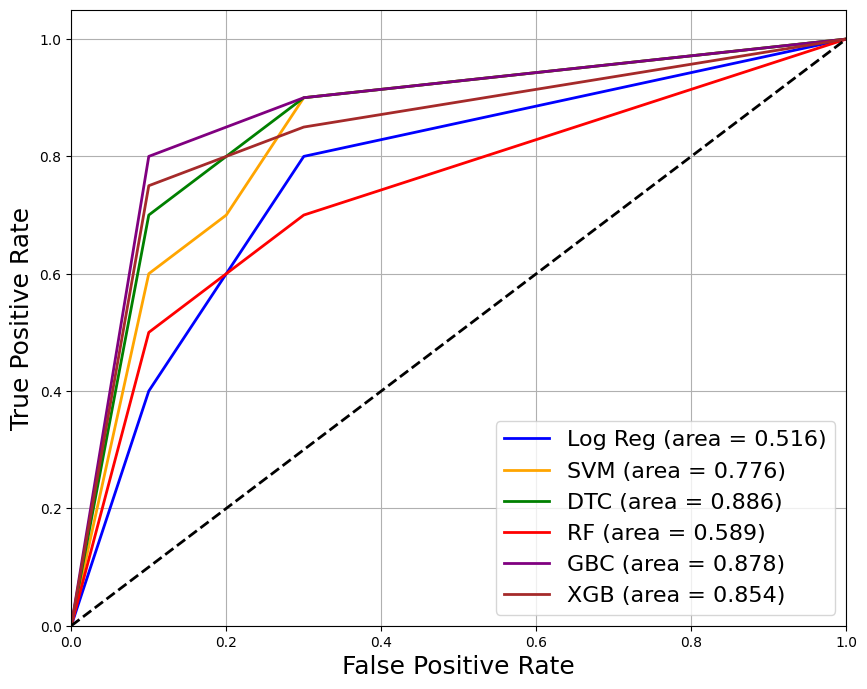}
    \caption{ROC curves for the first testing scenario of the imbalanced data}
    \label{fig 7}
\end{figure}

\begin{table}[h]
    \centering
    \caption{Summary results table for the first testing scenario of the imbalanced data}
    \label{tab:results_first_run}
    \begin{tabular}{cccccc}
        \toprule
        \textbf{Model} & \textbf{Balanced Accuracy} & \textbf{Precision} & \textbf{Recall} & \textbf{FAR} \\
        \midrule
        LR & 0.50 & 0.50 & 0.10 & 0.01 \\
        SVM & 0.71 & 0.50 & 0.59 & 0.04 \\
        DTC & 0.82 & 0.79 & 0.76 & 0.01 \\
        RF & 0.50 & 1.00 & 0.17 & 0.00 \\
        GBC & 0.83 & 0.66 & 0.79 & 0.04 \\
        XGB & 0.81 & 0.89 & 0.83 & 0.01 \\
        \bottomrule
    \end{tabular}
\end{table}

\subsubsection{Testing Scenario II: Oversampling SMOTE models}

After applying oversampling for the minority class by 70\%, it is observed that the use of SMOTE has improved performance by boosting AUC and recall for all models, especially for LR and RF. Because SMOTE generates a balanced training dataset, the models have more examples of training on, resulting in a better comprehension of the data distribution. There is a slight decrease in precision values with a corresponding increase in recall values and FAR. This is understandable due to the increase in the samples of minor classes, which allows the classifier to detect it more and misclassify the negative cases because of the trade-off between finding more defects and having fewer false alarms. As a result, models trained on balanced data may classify more errors and give a greater performance while increasing the number of false alarms. These findings highlight the significance of synthetic data production. Furthermore, the values of balanced accuracy remain stable for most models; only DTC shows a slight increase. Since the data is considerably balanced relative to the first run, the value of balanced accuracy demonstrates the actual accuracy of the models.

\begin{figure}[h]
    \centering
    \includegraphics[width=0.5\textwidth]{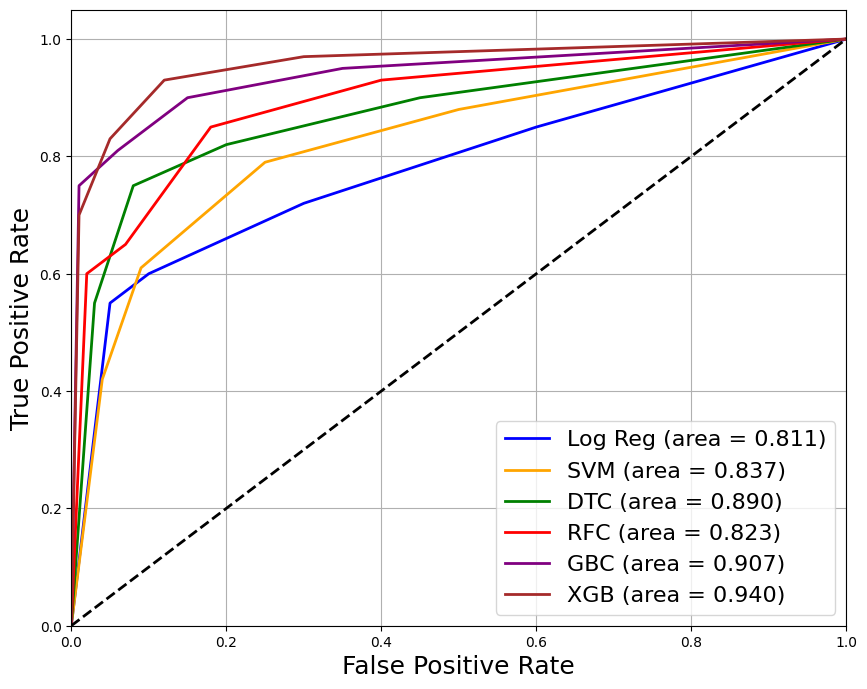}
    \caption{ROC curves for the second testing scenario of the SMOTE models}
    \label{fig 8}
\end{figure}

\begin{table}[h]
    \centering
    \caption{Summary results table for the second testing scenario of the SMOTE models}
    \label{tab:results_second_run}
    \begin{tabular}{cccccc}
        \toprule
        \textbf{Model} & \textbf{Balanced Accuracy} & \textbf{Precision} & \textbf{Recall} & \textbf{FAR} \\
        \midrule
        LR & 0.50 & 0.40 & 0.69 & 0.07 \\
        SVM & 0.55 & 0.49 & 0.72 & 0.05 \\
        DTC & 0.86 & 0.53 & 0.83 & 0.05 \\
        RF & 0.53 & 0.83 & 0.66 & 0.00 \\
        GBC & 0.83 & 0.80 & 0.83 & 0.01 \\
        XGB & 0.81 & 0.79 & 0.90 & 0.01 \\
        \bottomrule
    \end{tabular}
\end{table}

\subsubsection{Testing Scenario III: Combined resampling models}

The AUC and recall values substantially increase with combining under-sampling 80\% of the majority class and oversampling 40\% of the minority class, reaching 0.95 and 0.93 for XGB, respectively. This indicates a significant improvement in models with data resampling. In addition, the balanced accuracy of DTC increases to 88\%, which is considered the highest accuracy value. However, there is a noteworthy decrease in precision values and a slight increase in the values of FAR. The results slightly improved in this case due to the reduction in the synthetic data and the convergence of the observations of the two classes. The models accurately predicted the defect components. However, the balanced accuracy did not improve very well due to the small data size, which did not allow the model to be well-trained.

\begin{figure}[h]
    \centering
    \includegraphics[width=0.5\textwidth]{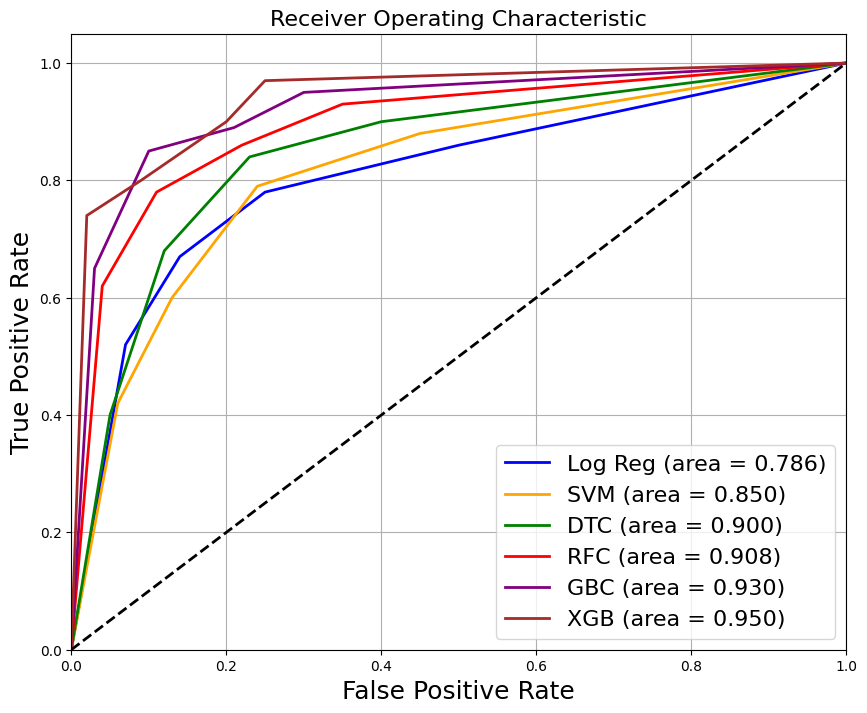}
    \caption{ROC curves for the third testing scenario of the combined resampling models}
    \label{fig 9}
\end{figure}

\begin{table}[h]
    \centering
    \caption{Summary results table for the third testing scenario of the combined resampling models}
    \label{tab:results_third_run}
    \begin{tabular}{cccccc}
        \toprule
        \textbf{Model} & \textbf{Balanced Accuracy} & \textbf{Precision} & \textbf{Recall} & \textbf{FAR} \\
        \midrule
        LR & 0.50 & 0.34 & 0.66 & 0.08 \\
        SVM & 0.55 & 0.46 & 0.76 & 0.06 \\
        DTC & 0.88 & 0.48 & 0.86 & 0.06 \\
        RF & 0.53 & 0.83 & 0.83 & 0.00 \\
        GBC & 0.83 & 0.62 & 0.90 & 0.04 \\
        XGB & 0.81 & 0.66 & 0.96 & 0.03 \\
        \bottomrule
    \end{tabular}
\end{table}

Finally, the results of the three runs are summarized in Figure 8 to demonstrate the trend of each performance metric. The balanced accuracy is not well-improved except for DT and LR, which considerably increase with resampling. The models show improvement in detecting the defects with resampling, which increased drastically to 93\% for XGB and 90\% for GBT. With the increase in detecting the errors, there is a slight corresponding decrease in classifying the conforming product, meaning an increase in the false alarm rate.

In summary, various performance metrics with different models are compared, and choosing the optimum and trained classifier will mainly depend on the model user's preferences. Many preferences tend to have a model that provides a balance between detecting the conforming products and defect ones with slight data imbalance as in the second and third run. Therefore, XGB with 95\% AUC will be the optimum choice. If the data is highly imbalanced as in the first run, it is better to consider the balanced accuracy, representing DTC with 88\%. Nevertheless, the main concern will be detecting the defects rather than the conforming products when handling expensive products. This means that the recall and precision metrics are important to consider; however, the recall will have a higher priority since recall focuses mainly on minimizing the number of positive cases, while precision focuses on minimizing the number of negative cases and positive cases. It can be noted from Figure 8 that resampling the data as in the third run resulted in a higher recall of 96\%, which inversely led to a decrease in the precision and an increase in the FAR. It is better to have a higher FAR with an increase in recall for an expensive manufacturing process. The results obtained in this study are compared with SECOM literature in Table \ref{tab:comparison_literature}.

\begin{table}[H]
    \centering
    \caption{Comparing the results with recent published journal papers}
    \label{tab:comparison_literature}
    \begin{tabular}{p{4cm} p{2cm} p{2.5cm} p{2.5cm} p{2.5cm}}
        \toprule
        \textbf{Performance Metric} & \textbf{Proposed Approach} & \textbf{Deep NN + PSO} \cite{lee2019data} & \textbf{In-painting KNN Imputation} \cite{kim2017particle}  & \textbf{Data-driven Approach} \cite{salem2018experimental} \\
        \midrule
        Precision & 0.66 & - & 0.16 & 0.80 \\
        Recall & 0.96 & 0.98 & 0.78 & 0.77 \\
        Balanced Accuracy & 0.83 & 0.86 (Accuracy) & - & 0.84 (Accuracy) \\
        AUC & 0.95 & - & 0.80 & - \\
        FAR & 0.03 & 0.16 & 0.09 & 0.25 \\
        \bottomrule
    \end{tabular}
\end{table}

\section{Conclusion}

In this work, the SECOM dataset, derived from a real-world semiconductor manufacturing plant, was thoroughly examined and classified. We evaluated 18 different approaches, incorporating various stages of data imputation, data imbalance handling, feature selection, and classification. Additionally, numerous trials were conducted, including selecting suitable algorithms for missing value imputation, hyperparameter tuning of models, and adjusting resampling percentages.

The proposed approach, which emphasizes rare-based feature selection and feature voting, demonstrated a significant improvement in model predictability for positive cases compared to existing methods in the literature. This approach effectively identified the most critical features, enhancing the model's ability to predict failures accurately. Moreover, the features with the highest voting could be further analyzed with additional sensor information to provide deeper insights into failure causes and identify the most crucial stages in the manufacturing process. This experimental evaluation identified the most suitable tools and stages for classifying the SECOM dataset. The results highlighted the superiority of XGB for classification, SMOTE for synthetic data generation, feature voting for feature selection, and mixed algorithms for missing data imputation. These findings underscore the effectiveness of our proposed methodologies in handling complex and imbalanced industrial datasets, paving the way for more reliable and accurate predictive models in semiconductor manufacturing.

As future work, leveraging Large Language Models (LLMs) and Generative AI could provide innovative solutions to address class imbalance issues \cite{ghali2024gamedx}. These advanced AI techniques can generate synthetic data and enhance data augmentation strategies, further improving model robustness and accuracy in handling imbalanced datasets.

%Bibliography

\end{document}